\documentclass{article}

\usepackage{arxiv}

\usepackage[utf8]{inputenc} 
\usepackage[T1]{fontenc}    
\usepackage{hyperref}       
\usepackage{url}            
\usepackage{booktabs}       
\usepackage{amsfonts}       
\usepackage{nicefrac}       
\usepackage{microtype}      
\usepackage{lipsum}		
\usepackage{graphicx}
\usepackage{hyperref}
\hypersetup{colorlinks,allcolors=black}
\usepackage{doi}
\usepackage{graphicx}
\usepackage[justification=centering]{caption}
\makeatletter
\newcommand*{\rom}[1]{\expandafter\@slowromancap\romannumeral #1@}
\makeatother
\usepackage{lineno}
\usepackage{float}
\usepackage{amsmath, amsthm, amssymb}
\newtheorem{prop}{Proposition}
\newtheorem{theorem}{Theorem}
\usepackage{lineno,hyperref}
\usepackage{siunitx}
\usepackage{makecell}
\usepackage{pbox}
\usepackage{fourier} 
\usepackage{array}
\usepackage{float}
\usepackage{lscape}
\usepackage[table,xcdraw]{xcolor}
\usepackage{physics}
\usepackage{breqn}
\usepackage{amsmath}
\usepackage{textcomp}
\usepackage{graphicx}
\usepackage[justification=centering]{caption}
\usepackage{lineno}
\usepackage{float}
\usepackage{amsmath,amssymb}
\usepackage{amsthm}
\usepackage{lineno,hyperref}
\usepackage{siunitx}
\usepackage{makecell}
\usepackage{pbox}
\usepackage{fourier} 
\usepackage{array}
\usepackage{xcolor}
\usepackage{float}
\usepackage{lscape}
\SetMathAlphabet{\mathtt}{normal}{OT1}{SourceCodePro-TLF}{m}{n}

\usepackage[table,xcdraw]{xcolor}

\title{Momentum Diminishes the Effect of Spectral Bias in Physics-Informed Neural Networks}

\author{ \href{}{\hspace{1mm}Ghazal Farhani}\\
	Department of Electrical and Computer Engineering\\
	The University of Western Ontario\\
	London, ON, Canada \\
	\texttt{gfarhani@uwo.ca} \\
	\And
\href{}{Alexander Kazachek} \\
	Department of Mathematics\\
	The University of Western Ontario\\
	London, ON, Canada \\
	\texttt{akazache@uwo.ca} \\
	\And
	\href{}{Boyu Wang} 
	\thanks{Author to whom correspondence should be addressed.} \\
	Department of Computer Science\\
	The University of Western Ontario\\
	London, ON, Canada \\
	\texttt{bwang@csd.uwo.ca } \\
}

\renewcommand{\shorttitle}{Momentum Diminishes the Effect of Spectral Bias in Physics-Informed Neural Networks}

\hypersetup{
pdftitle={A template for the arxiv style},
pdfsubject={q-bio.NC, q-bio.QM},
pdfauthor={Ghazal Farhani, Alexander Kazachek, Boyu Wang},
pdfkeywords={First keyword, Second keyword, More},
}

\begin{document}
\maketitle

\begin{abstract}
Physics-informed neural network (PINN) algorithms have shown promising results in solving a wide range of problems involving partial differential equations (PDEs). However, they often fail to converge to desirable solutions when the target function contains high-frequency features, due to a phenomenon known as spectral bias. In the present work, we exploit neural tangent kernels (NTKs) to investigate the training dynamics of PINNs evolving under stochastic gradient descent with momentum (SGDM). This demonstrates SGDM significantly reduces the effect of spectral bias. We have also examined why training a model via the Adam optimizer can accelerate the convergence while reducing the spectral bias. Moreover, our numerical experiments have confirmed that wide-enough networks using SGDM still converge to desirable solutions, even in the presence of high-frequency features. In fact, we show that the width of a network plays a critical role in convergence.
\end{abstract}


\section{Introduction}
Physics-informed neural networks (PINNs) have been proposed as alternatives to traditional numerical partial differential equations (PDEs) solvers \cite{raissi2019physics, raissi2020hidden, sirignano2018dgm, tripathy2018deep}. In PINNs, a PDE which describes the physical domain knowledge of a problem is added as a regularization term to an empirical loss function.  Although PINNs have shown remarkable performance in solving a wide range of problems in science and engineering \cite{cai2022physics, kharazmi2019variational, sun2020surrogate, kissas2020machine, tartakovsky2018learning}, they often fail to accurately construct solutions when the target function contains ``high-frequency'' features \cite{krishnapriyan2021characterizing, wang2021understanding}. This phenomenon exists in even the simplest linear PDEs. Since PINNs are relatively new, only a few studies have explored the potential causes of the convergence problems in these algorithms. Here, we briefly explain their approaches.

It has been shown that PINNs normally fail to converge to the solutions when the target functions have ``high-frequency'' features \cite{ krishnapriyan2021characterizing, wang2021understanding}. The issue in the training of the ``high-frequency''  features is not limited to PINNs. Rahman et. al \cite{rahaman2019spectral} empirically and Cao et. al \cite{cao2019towards} theoretically showed all fully-connected feed-forward neural networks (NNs) exhibit spectral bias, meaning NNs are biased to learn ``low-frequency'' features that correspond to less complex functions \cite{rahaman2019spectral}. To better understand the spectral bias of PINNs, by assuming an infinitely wide network, Wang et. al \cite{wang2022} employed neural tangent kernel (NTK) theory \cite{jacot2018neural} to analyze the training convergence of the networks. By examining the gradient flow of PINNs during the training, they showed that under the stochastic gradient descent (SGD) optimization process a very strong spectral bias was exhibited. Moreover, Wang et. al \cite{wang2021understanding} concluded that in the presence of high-frequency features the convergence to the solution under SGD may completely fail. They proposed to assign a weight to each term of the loss function and dynamically update it. Although the results showed some improvements, this approach could not completely minimize the effect of spectral bias. For target functions with higher frequency terms, their PINNs still failed to converge to the solution. Moreover, as assigning weights may result in indefinite kernels, the training process could become extremely unstable. 

Krishnapriyan et. al \cite{krishnapriyan2021characterizing} took a different approach. Instead of modifying the loss function terms, they implemented a curriculum learning method. In their approach, the model was gradually trained on target functions with lower frequencies, and, at each step, the optimized weights were used as the warm initialization for higher-frequency target functions. Their method showed good performance on some PDEs. However, as the frequency terms become larger, more steps are required and the step sizes between one frequency to the next one are not known. Thus, several trials and errors are required to find the optimized initialization terms. 

In the present work, we prove that an infinitely-wide PINN under the vanilla SGD optimization process will converge to the solution. However, for high-frequency terms the learning rate needs to become very small. This makes the convergence extremely slow, hence in practice not possible. Moreover, we prove that for infinitely-wide networks, using the SGD with momentum (SGDM) optimizer can reduce the effect of spectral bias in the networks, while significantly outperforming vanilla SGD. We also discuss why the Adam optimizer can accelerate the optimization process relative to SGD. Although the concept of infinitely-wide PINNs is not new, to our knowledge this is the first time that the gradient flow of the output of PINNs under the SGDM is being analyzed, and its relation to solving spectral bias is provided. Moreover, until now, infinitely-wide NNs were only used theoretically to facilitate the analysis of the dynamics of the learning process. In this work, by using the numerical experiments, we show that when dealing with high-frequency features, for fixed-depth networks, the training error decreases as the networks become wider. Hence, depending on the problem in hand, a sufficient width may be reached such that, for practical use, the training error becomes sufficiently small. 

%


\section{Preliminaries}
\subsection{PINNs General Form}
The general form of a well-posed PDE on a bounded domain ($\Omega \subset \textbf{R}^d$) is defined as:
\begin{equation}
\begin{aligned}
    \mathcal{D}[u](\textbf{\textit{x}}) &= f(\textbf{\textit{x}}) ,\; \; x \in \Omega \\
    u(\textbf{\textit{x}}) &= g(\textbf{\textit{x}}),\;  \; x \in  \partial \Omega
\end{aligned} 
\label{PDE_general}
\end{equation}
where $\mathcal{D}$ is a differential operator and $u(\textbf{\textit{x}})$ is the solution (of the PDE), in which $\textbf{\textit{x}} = (x_1, x_2, \dots, x_d)$. Note that for time-dependent equations, $\textbf{\textit{t}} = (t_1, t_2, \dots, t_d)$ are viewed as additional coordinates within $\textbf{\textit{x}}$. Hence, the initial condition is viewed as a special type of Dirichlet boundary condition and included in the second term. 

Using PINNs, the solution of Eq.~\ref{PDE_general} can be approximated as $u(\textbf{\textit{x}}, \textit{\textbf{w}})$. That is the approximation resulting from minimizing the following loss function:
\begin{equation}
    \mathcal{L}(\textit{\textbf{w}}) := \frac{1}{N_b} \sum_{i=1} ^{N_b} {(u({x_{b}}^i,\textit{\textbf{w}}) - g({x_{b}}^i))}^2 + \frac{1}{N_r} \sum_{i=1} ^{N_r} {(\mathcal{D}[u]({x_{r}}^i,\textit{\textbf{w}}) - f({x_{r}}^i))}^2
\label{loss_PDE}
\end{equation}
where $\{{x_{b}}^i\}_{i=1}^{N_b}$ and $\{{x_{r}}^i\}_{i=1}^{N_r}$ are boundary and collocation points respectively, and \textit{\textbf{w}} describes the neural network weights. The first term of Eq.~\ref{loss_PDE}, which we will denote as $\mathcal{L}_b(\textit{\textbf{w}})$, corresponds to the mean squared error of the boundary (and initial) condition data points. The second term, denoted $\mathcal{L}_r(\textit{\textbf{w}})$, encapsulates the physics of the problem using the randomly selected collocation points. Similar to all other NNs, minimizing the loss function $\mathcal{L}(\textit{\textbf{w}})$ results in finding the optimized solution for $u(\textbf{\textit{x}}, \textbf{\textit{w}})$. 

\subsection{Infinitely-Wide Neural Networks}
The output of a fully-connected infinitely-wide NN with $L$ hidden layers can be written as \cite{jacot2018neural}:
\begin{equation*}
\begin{split}
    \textbf{u}^{0}(\textbf{\textit{x}}) & = \textbf{\textit{x}}\\
    \textbf{u}^{h}(\textbf{\textit{x}}) &= \frac{1}{N}\mathtt{\Theta}^{h}\cdot\textbf{g}^{h} + \mathbf{b}^{h}  \\
    \textbf{g}^{h}(\textbf{\textit{x}}) &= \sigma(\mathtt{\Theta}^{h-1}\cdot\textbf{u}^{h-1}) + \mathbf{b}^{h-1}
\end{split}
\end{equation*}
where $\mathbf{\Theta}^h$ and $\textbf{b}^{h}$ are respectively the weight matrices and the bias vectors in the layers $h = 1, \dots, L$, $N$ is the width of the layer, $\textbf{\textit{x}}$ is the input vector, and $ \sigma(\cdot)$ is a $\beta$-smooth activation function (such as $\tanh(\cdot)$). The final output of the NN is written as:
\begin{equation*}
    \textbf{u}(\textbf{\textit{x}}, \textit{\textbf{w}}) = \textbf{f}^{L}(\textbf{\textit{x}}) = \frac{1}{N}\mathbf{\mathtt{\Theta}}^{L}.\textbf{g}^{L} + \textbf{b}^{L} 
\end{equation*}
where $\textit{\textbf{w}} = (\mathtt{\Theta}^0, \mathbf{b}^0, \dots,  \mathbf{\Theta}^L, \mathbf{b}^L)$. 

At each time step $t$ we can determine the change of the output with respect to the input, which defines a neural tangent kernel (NTK):
\begin{equation*}
    \textit{\textbf{K}}_{{(x,x')}}^t = \nabla_w{\textbf{u}(\textbf{\textit{x}}, \textit{\textbf{w(t)}})}^\top  \nabla_w{\textbf{u}(\textbf{\textit{x}}', \textit{\textbf{w(t)}})}.
\end{equation*}

One major consequence of dealing with infinitely-wide NNs is that if the last layer of the network is linear the NTK becomes static (does not change in time) \cite{liu2020linearity}. Another major consequence is that the norm of the Hessian of the output becomes smaller as the width become larger. In fact, as $N \to \infty$ the norm of the Hessian becomes $0$.  

\section{Theoretical Results}
\subsection{Convergence of the Gradient Descent Algorithm in the Presence of High-Frequency Features}{\label{theory_SGD}}

Generally, the optimization problems corresponding to over-parametrized systems, even on a local scale, are non-convex. A loss function $\mathcal{L}(\textit{\textbf{w}},x)$ of a $\mu$-uniformly conditioned NN satisfies the $\mu$-PL$_{*}$ condition on a set $S \subset \textbf{R}^m $ if:  
\begin{equation}
    \norm{\nabla_w(\mathcal{L}(\textit{\textbf{w}}))}^2 \geq \mu \mathcal{L}(\textit{\textbf{w}}), \forall \textit{\textbf{w}} \in S
\label{muPL}
\end{equation}
where $\mu$ is the lower bound of the tangent kernel $K(\textit{\textbf{w}})$ of the NN. 
It has been shown that infinitely-wide networks satisfy the $\mu$-PL$_{*}$ condition, a variant of the Polyak-Lojasiewicz condition, and as a result the (stochastic) gradient descent (SGD/GD) optimization algorithms will converge to the optimal solution \cite{liu2022loss}. The following proposition makes use of the $\mu$-PL$_{*}$ condition to provide a convergence analysis for an infinitely-wide PINN with a loss function as in Eq.~\ref{loss_PDE} optimized with SGD (Appendix~\ref{PINN_converge}).


\begin{prop}
Let $\lambda_{b_\max}$ and $\lambda_{r_\max}$, respectively, be the largest eigenvalues of the Hessians $\nabla^2\mathcal{L}_b(\textit{\textbf{w}}^t)$ and  $\nabla^2\mathcal{L}_r(\textit{\textbf{w}}^t)$. Consider an infinitely-wide PINN optimized with the following update rule:
\begin{equation*}
    \textit{\textbf{w}}^{t + 1} = \textit{\textbf{w}}^t - \eta \nabla{\mathcal{L}(\textit{\textbf{w}}^t)}
\end{equation*}
where $\eta$ is a constant learning rate. Then, provided $\eta=\mathcal{O}(1/(\lambda_{b_\max}+\lambda_{r_\max}))$, the $\mu$-PL$_{*}$ condition is satisfied and the PINN will converge.
\end{prop}

Unfortunately, for PDEs exhibiting stiff dynamics (those with large parameters) the eigenvalues (of the Hessian) dictating convergence are often very large. For example, take the one-dimensional Poisson equation:
\begin{equation}
\begin{split}
    u_{xx}(\textbf{\textit{x}}) &= s(\textbf{\textit{x}}), \textbf{\textit{x}} \in \mathtt{\Omega} \\
    u(\textbf{\textit{x}}) &= b(\textbf{\textit{x}}), \textbf{\textit{x}} \in  \partial \mathtt{\Omega}
\label{general_poission}
\end{split}
\end{equation}
with $u(\textbf{\textit{x}}) = \sin(C\textbf{\textit{x}})$. The Hessian of loss function $\mathcal{H}(\textit{\textbf{w}}^t)$ is of order $\mathcal{O}(C^4)$ (Appendix~\ref{poisson_stiff_dynamics}). Thus as $C$ grows so does the bound on the eigenvalues of $\mathcal{H}(\textit{\textbf{w}}^t)$, indicating that at least one of $\lambda_{b_\max}$ or $\lambda_{r_\max}$ will be large. This makes the learning rate necessary in order to guarantee the convergence of the PINN prohibitively small, and therefore SGD is unusable in practice.

  
\subsection{Wide PINN Optimization using SGD with Momentum}

In order to accelerate learning, we go beyond the vanilla gradient descent and investigate the  training dynamics of PINNs under SGD with momentum (SGDM)~\cite{du2019frontier}:
\begin{equation}\label{sgdm}
    \textit{\textbf{w}}_{t+1} = \textit{\textbf{w}}_{t} + \alpha(\textit{\textbf{w}}_{t} - \textit{\textbf{w}}_{t-1}) - \eta \nabla_{w}\mathcal{L}(\textit{\textbf{w}})
\end{equation}
where \(\alpha\) and \(\eta\) are fixed learning rates.

Specifically, we analyze the gradient flow of the update rule in Eq.~\ref{sgdm} by leveraging the notion of neural tangent kernels (NTK). The following theorem (proof in Appendix~\ref{SGDM_flow}) reveals that SGDM has a different convergence behavior from SGD:
\begin{theorem}
For an infinitely-wide PINN, the gradient flow of SGDM is:
\begin{equation*}
m \begin{bmatrix}
\ddot{u}(\textit{\textbf{x}}_{b},\textit{\textbf{w(t)}})   
\\
\textit{D}[\ddot{u}](\textit{\textbf{x}}_{r},\textit{\textbf{w(t)}})
\end{bmatrix} = -\mu \begin{bmatrix}
\dot{u}(\textit{\textbf{x}}_{b},\textit{\textbf{w(t)}})   
\\
\textit{D}[\dot{u}](\textit{\textbf{x}}_{r},\textit{\textbf{w(t)}}) \end{bmatrix} -
\textit{\textbf{K}} \begin{bmatrix}
u(\textit{\textbf{x}}_{b},\textit{\textbf{w(t)}}) - g(\textit{\textbf{x}}_{b}) \\
\textit{D}[u](\textit{\textbf{x}}_{r},\textit{\textbf{w(t)}}) - h(\textit{\textbf{x}}_{r})
\end{bmatrix}
\end{equation*}
where $\mu$ is related to momentum by: $\alpha = \frac{m}{m+\mu \Delta t}$ and $\textit{\textbf{K}}$ is: 
\begin{equation*}
\textit{\textbf{K}} = 
\begin{bmatrix}
\textit{\textbf{K}}_{{bb}}  & \textit{\textbf{K}}_{{rb}} \\
\textit{\textbf{K}}_{{br}} & \textit{\textbf{K}}_{{rr}}
\end{bmatrix},
\end{equation*}
and:
\begin{equation*} 
\begin{split}
\textit{\textbf{K}}_{{bb}_{(x,x')}} & = \nabla_w{u(\textit{\textbf{w}},\textit{\textbf{x}})}^\top  \nabla_w{u(\textit{\textbf{w}},\textit{\textbf{x}}')} \\
\textit{\textbf{K}}_{{br}_{(x,x')}} & = \nabla_w{u(\textit{\textbf{w}},\textit{\textbf{x}})}^\top  \nabla_w{\textit{D}[u](\textit{\textbf{w}},\textit{\textbf{x}}')} \\
\textit{\textbf{K}}_{{rr}_{(x,x')}} & = \nabla_w{\textit{D}[u](\textit{\textbf{w}},\textit{\textbf{x}}')}^\top  \nabla_w{\textit{D}[u](\textit{\textbf{w}},\textit{\textbf{x}}')}.
\label{kernel}   
\end{split}
\end{equation*}
are  three NTKs associated to the boundary and residual terms. Moreover, let  \(\gamma=\mu/2m\), \(\kappa_i\) be the eigenvalues of \(\textit{\textbf{K}}\), and $\kappa'_{i} = \frac{\kappa_{i}}{m}$. Then, the solutions to the gradient flow are of the form:
\begin{equation}
\begin{split}
    A_1 e^{ \lambda_{{i}_1} t} +  A_2 e^{ \lambda_{{i}_2} t}\\
    \lambda_{{i}_{1,2}} = -\gamma \pm \sqrt{\gamma^2 - \kappa'_{i}}
\end{split}
\label{damping-OCS}
\end{equation}
where \(A_1\) and \(A_2\) are constants.
\end{theorem}

Examining the analogous gradient flow for the vanilla SGD, the training error decays at the rate of $ e ^{-\kappa_{i} t}$ \cite{wang2021understanding}. That is, the eigenvalues of the kernel indicate how fast the training error of SGD decreases. Spectral bias arises as eigenvectors of larger eigenvalues of the kernel matrix represent lower frequencies \cite{rahaman2019spectral, basri2020frequency}. Thus, it is clear that under vanilla SGD the high-frequency components are extremely slow to converge. 

\begin{figure}[h]
\centering
\includegraphics[width=0.5\columnwidth]{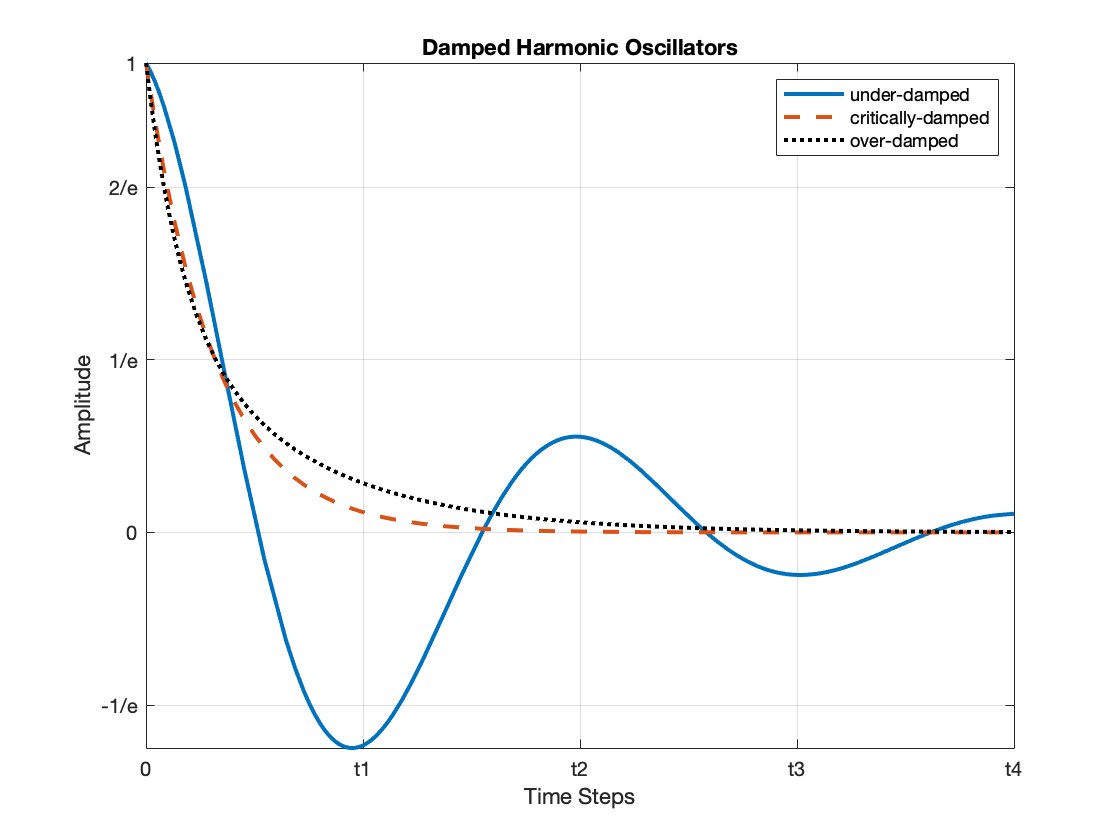}
\caption{Damped harmonic oscillators show three different characteristics.}
\label{fig:damped_OSC}
\end{figure}

Once we add momentum, the decay rate analysis becomes more involved as Eq.~\ref{damping-OCS} yields three different cases of solutions. Each of the three cases are analogous to one of the solutions of a damped harmonic oscillator in physics~\cite{qian1999momentum, aryaintroduction} (Fig.~\ref{fig:damped_OSC}).
\begin{itemize}
    \item Under-damped: Imaginary roots ($\gamma^2 < \kappa'_i$)
    \item Critically-damped: Real and equal roots ($\gamma^2 = \kappa'_i)$
    \item Over-damped: Real roots ($\gamma^2 > \kappa'_i$).
\end{itemize}


{\bf Under-damped case } As $\sqrt{\gamma^2 - \kappa_i}$ has imaginary roots, Eq.~\ref{damping-OCS} can be rewritten as: 
\begin{equation*}
A e^{-\gamma t} \cos{(\omega_{1} t + \phi)} \\
\label{under-damping}
\end{equation*}
where $\omega_{1} = \sqrt{\kappa_i- \gamma^2}$, and $A$ and $\phi$ are two constants corresponding to the amplitude and the phase of the damped oscillation. In physics, this solution corresponds to an oscillatory motion in which the amplitude is decaying exponentially. 


{\bf Critically-damped case }  The general solution of the critically-damped case can be written as: 
\begin{equation*}
(B_1 + B_2t) e^{-\gamma t}
\label{critically-damping}
\end{equation*}
where $B_1$ and $B_2$ are constants. As the oscillation motion is not present, the decaying rate for this case is much faster (Fig.~\ref{fig:damped_OSC}). 

{\bf Over-damped case }  Lastly, in the over-damped case the general solution is simplified as: 
\begin{equation*}
e^{-\gamma t} \left (C_1 e^{\omega_{2}} t + C_2 e^{-\omega_{2}} t\right)
\label{over-damping}
\end{equation*}
where $\omega_{2} = \sqrt{\gamma^2 - \kappa_i}$ and $C_1$ and $C_2$ are constants. Similar to the critically-damped case, the above equation states a fast decay. 

Thus, depending on the absolute value of the eigenvalues of the kernel matrix ($\abs{\lambda_i}$), the dynamics of the training error can differ. For larger eigenvalues the training error corresponds to an under-damped solution, in which the amplitude of an oscillatory motion is decaying exponentially, whereas for smaller eigenvalues the correspondence is with an over-damped or critically-damped oscillation, with much faster decay. Thus, when using SGDM instead of vanilla SGD, the training process for high-frequency components of the target (corresponding to the smaller eigenvalues) will decay faster (undergoing the over-damped or critically-damped motions) while the low-frequency components (corresponding to the larger eigenvalues) will decay at a slower rate. As a consequence, the effect of spectral bias will be less prominent (compared to vanilla SGD). Of note, the fastest decay is observed for the critically-damped case, in practice many small eigenvalues are near the value of $\gamma$ and the dynamics of the training process becomes very close to the critically-damped case. 

\subsection{Analysis of Adam for Band-Limited Functions}

Adam is a commonly-used optimizer which can be interpreted as SGDM adapted for variance \cite{kingma2014adam,balles2018dissecting}. Optimizing with any gradient-descent-based method, such as Adam, we sample a batch \(X\) of \(N\) data points, and compute the stochastic gradient \(g(\textit{\textbf{w}};X)=\sum_{\textit{\textbf{x}}\in X}\nabla\ell(\textit{\textbf{w}};\textit{\textbf{x}})/N\) (where in our case \(\ell\) is the respective loss for whether \(\textit{\textbf{x}}\) is a boundary or collocation point). Define the pointwise variance \(\sigma=\operatorname{var} g(\textit{\textbf{w}};X)\), and denote by \(\sigma_i\) and \(\nabla\mathcal{L}_i\) their \(i\)-th entries (henceforth we stop writing \(\textit{\textbf{w}}\) for clarity). At each epoch, Adam then updates the \(i\)-th entry of \(\textit{\textbf{w}}\) with a magnitude inversely proportional to an estimator of \(\sigma_i^2/\nabla\mathcal{L}_i^2\) \cite{balles2018dissecting}. The end-goal of the training process is to find a \(\textit{\textbf{w}}\) which minimizes the expected loss across all collocation and boundary points \(L\), approximated by \(\mathcal{L}\) in Eq.~\ref{loss_PDE}.

Our numerical experiments confirm that for wide NNs both SGDM and Adam can converge to desirable solutions, but that Adam is even faster than SGDM (see Section~\ref{sec:Numerical}). This may be due to variance adaptation alleviating spectral bias even further. The connection is readily seen when the solution to our PDE \(u(\textit{\textbf{x}})\) is (at least well-approximated by) a band-limited function, meaning \(u(\textit{\textbf{x}})=\sum_{k\in K} \alpha_k\exp(2\pi i k\textit{\textbf{x}})\) for some finite set of frequencies \(K\). 

For such solutions we eventually satisfy the bound:
\begin{equation*}
    \mathcal{L}\approx L\leq \sqrt{\frac{2\pi \sum_{k\in K}\alpha_k^2 k^2}{N}}
\end{equation*}
with high probability \cite{basri2019frequency,arora2019overparametrized}. This indicates the loss is only guaranteed to be small if the \(k\in K\) are themselves small, precisely when \(u(\textit{\textbf{x}})\) consists of low frequencies. In other words, \(\mathcal{L}\) can be very large for high-frequency solutions. Since our NN is wide and thus satisfies the \(\mu\)-PL\(_\ast\) condition \(\|\nabla\mathcal{L}\|^2\geq\mu\mathcal{L}\) of Eq.~\ref{muPL}, such solutions imply \(\|\nabla\mathcal{L}\|\) and thus the \(\nabla\mathcal{L}_i\) also have loose bounds. 

Convergence happens when \(\nabla\mathcal{L}_i\approx 0\) for all \(i\), so if any \(\nabla\mathcal{L}_i\gg 0\) the NN is far away from the solution. However, this is precisely the situation in which Adam takes large steps and accelerates toward the solution fastest, as \(\sigma_i^2/\nabla\mathcal{L}_i^2\) is close to zero. This phenomenon certainly gives Adam an edge over SGD, and may contribute to it outperforming SGDM too.



\section{Numerical Experiments}\label{sec:Numerical}
In this section, using Poisson's equation, transport function, and the reaction-diffusion problem we provide results from our numerical experiments. 

\subsection{Poisson's Equation}
Poisson's equation is a well-known elliptic PDE in physics. For example, in electromagnetism, the solution to Poisson's equation is the potential field of a given electric charge. In Eq.~\ref{general_poission} the general form of the one-dimensional Poisson's equation was presented. Here, we write the Poisson's equation for a specific source function and a specific boundary condition:
\begin{align*}
    f(x) &= -C^{2} \sin{Cx},  \; \; x \in [0,1] \\
    g(x) &= 0, \; \; x = 0,1
\end{align*}
with $u(x) = \sin(Cx)$ as the solution. The corresponding loss function is written as:
\begin{equation*}
    \mathcal{L}(\textit{\textbf{w}}) := \frac{1}{N_b} \sum_{i=1} ^{N_b} {(\hat{u}({x_{b}}^i,\textit{\textbf{w}}) - g({x_{b}}^i))}^2 + \frac{1}{N_r} \sum_{i=1} ^{N_r} {(\hat{u}_{xx}({x_{r}}^i,\textit{\textbf{w}}) - f({x_{r}}^i))}^2
\end{equation*}
where $\hat{u}$ is the output of the network. We chose $N_r = 1000$ and $N_b = 1000$ and built a simple NN with 1 hidden layer of width 500. As expected, when $C = 5\pi$, after only $45000$ epochs, trained models via all three algorithms could accurately estimate the solution. The relative error $\norm{(u - \hat{u})/\hat{u}}$ of the output of NN using the vanilla SGD was on order of $10^{-2}$, indicating that the model could successfully predict the solution for the PDE. The relative error using SGDM and Adam were respectively on order of $10^{-3}$ and $10^{-4}$. The plots of estimated and exact solutions as well as the point-wise error for the mentioned algorithms are presented in Fig.~\ref{fig:Poission_C5}.  

\begin{figure}[h]
\centering
\includegraphics[width=0.9\columnwidth]{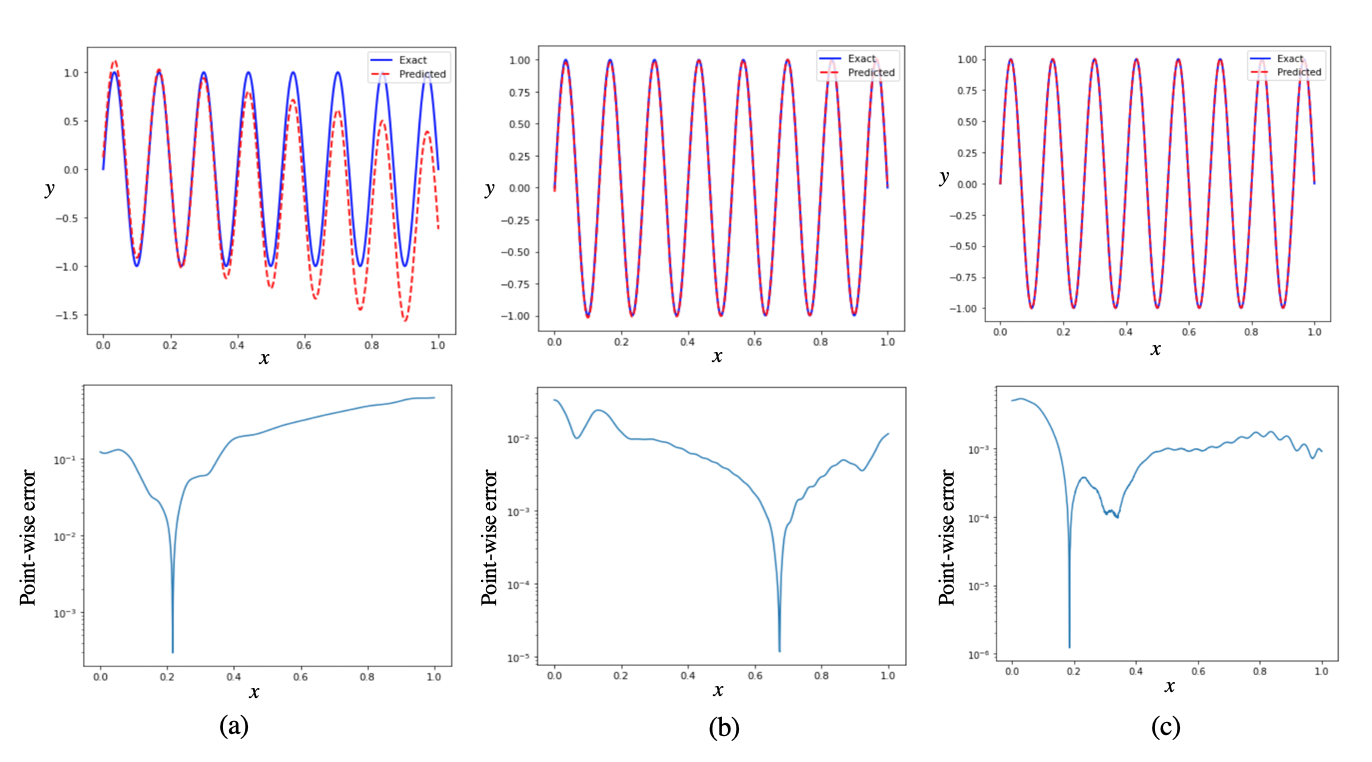}
\caption{1D Poisson equation when $C = 15 \pi$: (a) Vanilla SGD after with relative error on order of $10^{-1}$(b) SGDM after $40000$ epochs with relative error on order of $10^{-2}$. (c) Adam with relative error on order of $10^{-3}$.}
\label{fig:Poission_C15}
\end{figure}

As the parameters of a PDE grow larger, converging to the solution becomes more challenging. For Poisson's equation when $C = 15\pi$, after $55000$ epochs, the obtained solution from training the network with the vanilla SGD exhibited a relative error on order of $10^{-1}$. Meanwhile the trained model with SGDM, after being trained for $55000$ epochs, provided an acceptable solution with a relative error on order of $10^{-2}$. Moreover, Adam converged to the solution after training the model for only $25000$ epochs. The relative error was also considerably smaller and on order of $10^{-3}$. The comparison between the results of training the model using the three algorithms is shown in Fig.~\ref{fig:Poission_C15}. Of note, using vanilla SGD and training the model for $180000$ epochs resulted in a small relative error on the order of $10^{-2}$ (Fig.~\ref{fig:poission_18K_SGD}). This confirmed our discussion presented in Section \ref{theory_SGD} that implementing vanilla SGD will result in convergence under large parameters, though due to the presence of high-frequency features, the convergence becomes extremely slow. 

\subsection{Transport Equation}
\begin{figure}[h]
\centering
\includegraphics[width=0.77\columnwidth]{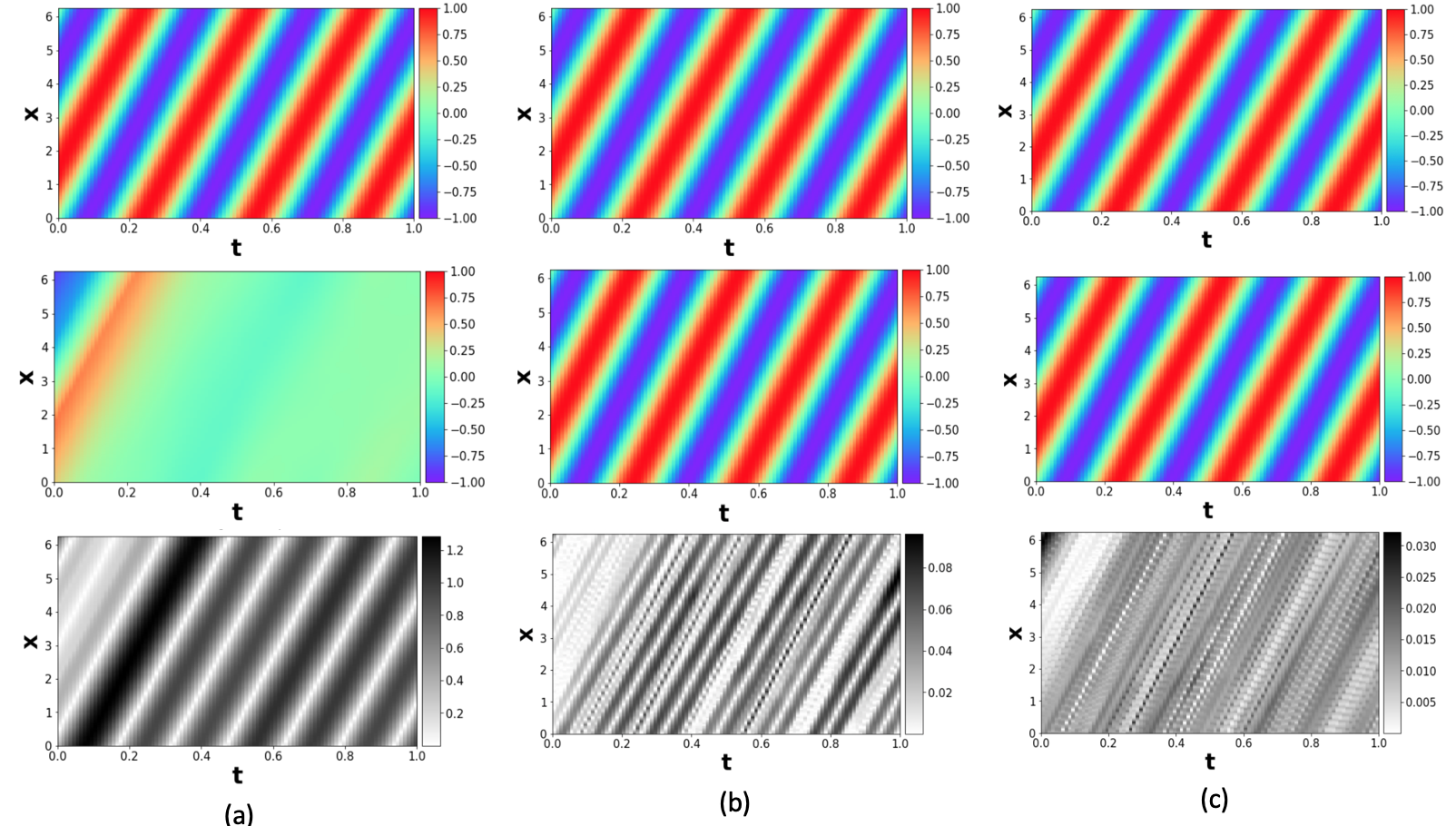}
\caption{1D transport equation when $\beta = 20$: The solutions are obtained by training a $4$-layer network with width=$500$ at each layer. Top panels: The exact (analytical) solution. Middle panels: The estimated solutions. Bottom panels: The absolute difference between the exact and estimated solutions. (a) Vanilla SGD, (b) SGDM ,(c) Adam.}
\label{fig:transport_beta20}
\end{figure}

The transport equation is a hyperbolic PDE that models the concentration of a substance flowing in a fluid. Here, we focus on a one-dimensional transport equation: 
\begin{equation*}
\begin{split}
    u_t + \beta u_x &= 0,  \; \; x \in \mathtt{\Omega}, T \in [0,1]\\
    g(x) &= u(x,0),   \; \; x \in \mathtt{\Omega}    
\end{split}
\end{equation*}
where $g(x)$ is the initial condition, and $\beta$ is a constant parameter. Using the methods of characteristics, the transport function has a well-defined analytical solution: $u(x,t) = g(x-\beta t)$. The corresponding loss function is written as:
\begin{equation*}
    \mathcal{L}(\textit{\textbf{w}}) := \frac{1}{N_b} \sum_{i=1} ^{N_b} {(\hat{u}({x_{b}}^i,{t_{b}}^i,\textit{\textbf{w}}) - g({x_{b}}^i))}^2 + \frac{1}{N_r} \sum_{i=1} ^{N_r} {(\hat{u}_t({x_{r}}^i,{t_{b}}^i,\textit{\textbf{w}}) + \beta \hat{u}_x({x_{r}}^i,{t_{b}}^i,\textit{\textbf{w}}))}^2
\label{loss_PDE_transort}
\end{equation*}
where $\hat{u}$ is the output of the network. We used $N_b = 450$ and $N_r = 4500$. We also chose the boundary and initial conditions to be:
\begin{equation*}
\begin{split}
    u(x, 0) &= \sin(x)\\
    u(0, t) &= u(2\pi, t).
\end{split}
\end{equation*}
We trained a 4-layer network with width = $500$ (at each layer). Similar to the Poisson's equation, for small $\beta$ values, the trained models with vanilla SGD, SGDM, and Adam optimization algorithms could easily converge to the solution, and had small relative errors. However, for $\beta = 20$, after $125000$ epochs the model trained with vanilla SGD optimizer still failed to find the solution, and the averaged relative error stood at a large value on order of $10^{0}$. However, the model trained via SGDM after 55000 epochs could converge to the solution and the estimated solution had the relative error on order of $10^{-2}$. The estimated solution from training the model via Adam, after only 15000 epochs, had a small relative error on the order of $10^{-2}$. The exact solution, the estimated solutions (based on the three optimizers), and the absolute difference between the exact and estimated solutions are shown in Fig.~\ref{fig:transport_beta20}. 

\subsection{Reaction-Diffusion Equation} \label{reaction-diffusion}
\begin{figure}[h]
\centering
\includegraphics[width=0.8\columnwidth]{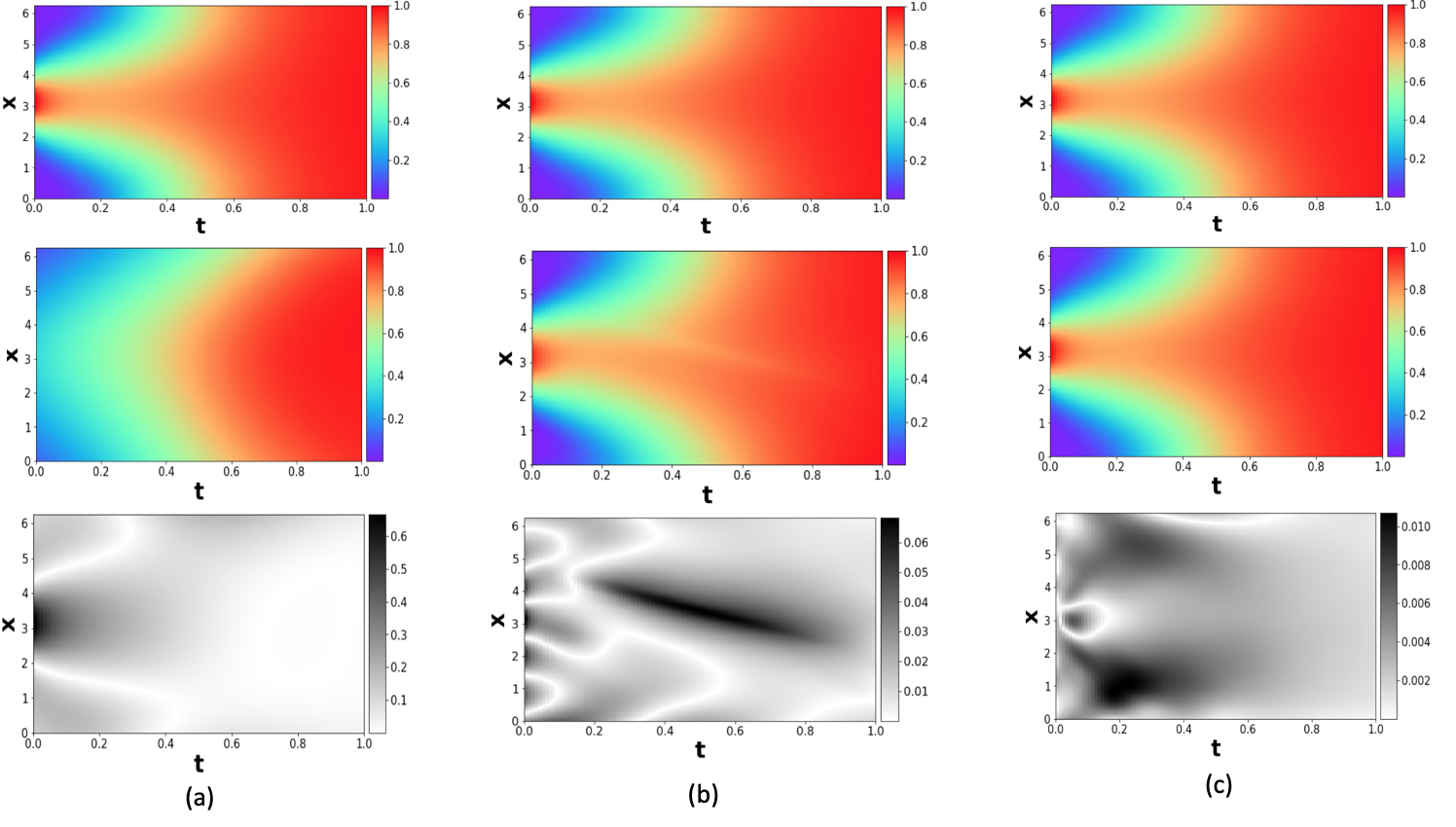}
\caption{1D reaction-diffusion equation for $\nu = 3$ and $\rho = 5$. The solutions are obtained by training a $4$-layer network with width=$500$ at each layer. Top panels: The exact (analytical) solution. Middle panels: The estimated solution. Bottom panels: The absolute difference between the exact and estimated solutions. (a) Vanilla SGD, (b) SGDM, (c) Adam. }
\label{fig:reaction}
\end{figure}

A reaction-diffusion equation contains a reaction and a diffusion term. Its general form is written as: 
\begin{equation*}
    u_{t} = \nu  \mathtt{\Delta} u + f(u)
\label{equ-reaction-diffusion}    
\end{equation*}
where $u(\textbf{\textit{x}},\textbf{\textit{t}})$ is the solution describing the density/concentration of a substance, $\mathtt{\Delta}$ is the Laplace operator, $\nu$ is a diffusion coefficient, and $f(u)$ is a smooth function describing processes that change the present state of $u$ (for example, birth, death or a chemical reaction). Here, we assume a one-dimensional equation, where $f(u) = \rho u (1-u)$. For a constant value of $\rho$ and with the defined $f(u)$, Equ.~\ref{equ-reaction-diffusion}, can be solved analytically \cite{krishnapriyan2021characterizing}. To estimate the solution, using PINNs, the loss function for the 1D reaction-diffusion PDE is written as: 
\begin{multline*}
        \mathcal{L}(\textit{\textbf{w}}) :=  \frac{1}{N_b} \sum_{i=1} ^{N_b} (\hat{u}(x_{b}^i,t_{b}^i,\textit{\textbf{w}}) - g(x_{b}^i))^2 \\
        + \frac{1}{N_r}\sum_{i=1}^{N_r}\left(\hat{u}_t(x_{r}^i,t_{b}^i,\textit{\textbf{w}}) - \nu \hat{u}_{xx}(x_{r}^i,t_{b}^i,\textit{\textbf{w}}) - \rho \hat{u}(x_{r}^i,t_{b}^i) (1-\hat{u}(x_{r}^i,t_{b}^i))\right)^2.
\label{loss_PDE_transort}
\end{multline*}
We further assigned $N_r = 1000$, and $N_n = 100$. We also chose the initial and boundary conditions as $u(x, 0) =  \exp {-\frac{(x-\pi)^{2}}{\frac{\pi^{2}}{2}}}$, and $u(0, t) = u(2\pi, t)$ respectively. Similar to the previous experiments, for larger choices of $\nu$ and $\rho$ the model trained via vanilla SGD (after $85000$ epochs) had difficulty to converge to the solution. However, models trained via SGDM and Adam (after $45000$ epochs) could provide solutions with small errors. The plots of estimated and exact solutions for the three algorithms when $\nu = 3$ and $\rho = 5$ are shown in Fig.~\ref{fig:reaction}.
 
\subsection{Width of Neural Networks}\label{Sec:width}
\begin{figure}[h]
\centering
\includegraphics[width=0.9\columnwidth]{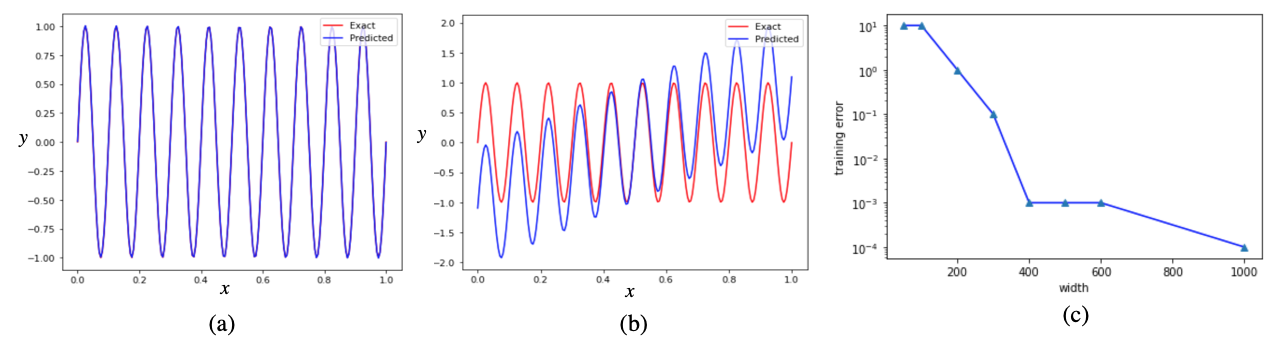}
\caption{Poisson's equation when C = 20$\pi$. (a) The solution is obtained by training a $1$-layer network. The width of each layer equals 50. (b) The solution is obtained by training an $8$-layer network. The width of each layer equals 50. }
\label{fig:width-analysis1}
\end{figure}

\begin{figure}[h]
\centering
\includegraphics[width=0.8\columnwidth]{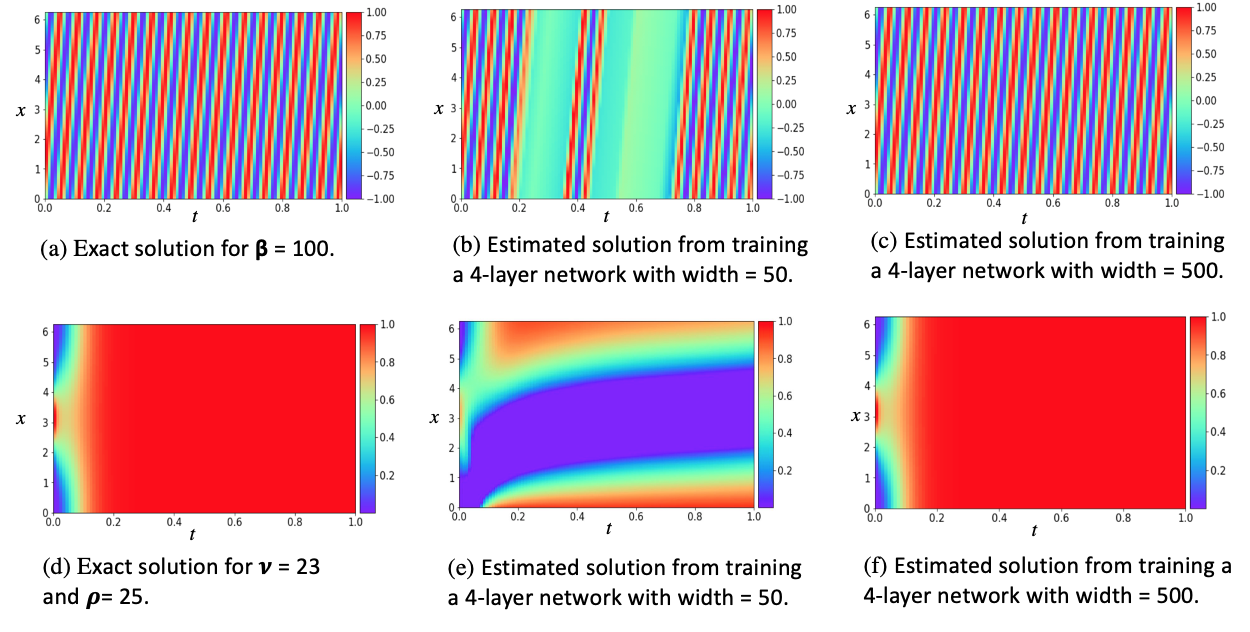}
\caption{Top panel: transport function when $\beta$ = 100. Bottom panel: The reaction-diffusion problem when $\nu$ = 23 and $\rho$ = 25.}
\label{fig:width-analysis2}
\end{figure}

An interesting observation was that for PDEs with large parameters the width of the neural networks played an important role. For example, in Poisson's equation with $C = 20\pi$, we used an $8$-layer NN with a width of $50$ neurons (per layer), and trained the model via Adam optimizer. The final output had a large relative error of order of $10^{0}$, while a shallow network with a width of $500$ neurons could converge to the solution and had small error on order of $10^{-3}$ (Fig.~\ref{fig:width-analysis1}(a) and (b)). Moreover, for a fixed depth of $2$, we trained several networks at different widths. As the networks became wider the relative error decreased (Fig.~\ref{fig:width-analysis1}(c)). Also, in the transport function, for $\beta = 100$, a $4$-layer network with a width of $50$ had a large relative error on order of $10^{-1}$. However, another $4$-layer network with a width of $500$ neurons had a considerably smaller relative error on order of $10^{-2}$. Similarly, in the reaction-diffusion equation for $\nu = 23$ and $\rho = 25$, a $4$-layer network with a width of $50$ had a large relative error on order of $10^{-1}$, while training another $4$-layer network with a width of $500$ resulted in a much smaller error on order of $10^{-3}$ (Fig.~\ref{fig:width-analysis2}).  

\vspace{-1em}
\section{Conclusion}
In the present study, through the lens of NTKs, we examined the dynamics of training PINNs via SGDM. We also showed that under SGDM the conversion rate of low-frequency features becomes slower (analogous to the under-damped oscillation) and many low-frequency features undergo much faster conversion dynamics (analogous to over-damped and critically-damped oscillations). Thus, the effect of spectral bias becomes less prominent. Moreover, we discussed how training a PINN via Adam can also accelerate the convergence, and showed it can be much faster than SGDM. Although we analyzed the dynamics of convergence by assuming an extreme case of infinitely-wide networks, our experiments confirmed that even when the width of the networks are not infinite, the estimated solutions obtained from the trained models via SGDM and Adam had high accuracy. Moreover, we observed that when dealing with target functions with high-frequency features, wider networks (compared to their narrower counterparts) could estimate solutions with much smaller errors. The fact that the width of the networks significantly enhanced the performance of the models indicates that the neural architecture may play an important role in PINNs. Thus, we are planning to explore more neural architecture choices. Moreover, one major future step is to examine if PINNs can solve more complicated PDEs. For example, researching if they are capable of solving highly non-linear equations, and if spectral bias becomes a more serious problem for those PDEs.

\bibliographystyle{unsrt}
{\footnotesize
\bibliography{PINN.bib}}

\appendix

\section{Convergence of Wide PINNs} \label{PINN_converge}

The loss function at time $t+1$ can be written as: 
\begin{equation*}
    \mathcal{L}(\textit{\textbf{w}}^{t+1}) \approx \mathcal{L}(\textit{\textbf{w}}^{t}) + (\textit{\textbf{w}}^{t + 1} - \textit{\textbf{w}}^t)^{\top} \nabla{\mathcal{L}(\textit{\textbf{w}}^t)} + \frac{1}{2} (\textit{\textbf{w}}^{t + 1} - \textit{\textbf{w}}^t)^{\top} \mathcal{H}(\textit{\textbf{w}}^t) (\textit{\textbf{w}}^{t + 1} - \textit{\textbf{w}}^t) 
\end{equation*}
where $\mathcal{H}(\textit{\textbf{w}}^t) = \nabla^2{\mathcal{L}_b (\textit{\textbf{w}}^t)} + \nabla^2{\mathcal{L}_r (\textit{\textbf{w}}^t)}$. Following the approach of \cite{wang2021understanding}, let $\textit{\textbf{Q}}$ be an orthogonal matrix diagonalizing $\mathcal{H}(\textit{\textbf{w}}^t)$ and $\textit{\textbf{v}}=  \mathcal{L}(\textit{\textbf{w}}^t)/\|\mathcal{L}(\textit{\textbf{w}}^t)\|$. With $\textit{\textbf{y}}=\textit{\textbf{Q}}\textit{\textbf{v}}$, we have the following:
\begin{align*}
    \mathcal{L}(\textit{\textbf{w}}^{t+1}) &\approx \mathcal{L}(\textit{\textbf{w}}^{t}) - \eta \norm{\nabla{L(\textit{\textbf{w}}^t)}} ^2 + \frac{\eta ^2}{2} \norm{\nabla{\mathcal{L}(\textit{\textbf{w}}^t)}} ^2 \left(\sum_i ^{N_b} \lambda_{b_{i}} {y_i}^{2} + \sum_i ^{N_r} \lambda_{r_{i}} {y_i}^{2}\right) \\
     &\lessapprox \mathcal{L}(\textit{\textbf{w}}^{t}) - \eta \norm{\nabla{\mathcal{L}(\textit{\textbf{w}}^t)}} ^2 + \frac{\eta ^2}{2} \norm{\nabla{\mathcal{L}(\textit{\textbf{w}}^t)}} ^2 (\lambda_{b_\max} + \lambda_{r_\max})
\end{align*}
where the $\lambda_{b_i}$ and $\lambda_{r_i}$ are the respective eigenvalues of the Hessians of $\mathcal{L}_b$ and $\mathcal{L}_r$ ordered non-decreasingly, and the summation is taken over the components of $\textit{\textbf{y}}$.

From here, fixing any bound $\mathcal{B}$ such that $\lambda_{b_\max} + \lambda_{r_\max}\leq\mathcal{B}$, we obtain the following inequality:
\begin{equation*}
    \mathcal{L}(\textit{\textbf{w}}^{t+1}) \lessapprox \mathcal{L}(\textit{\textbf{w}}^{t}) - \eta \norm{\nabla{\mathcal{L}(\textit{\textbf{w}}^t)}} ^2 \left(1 - \frac{\eta \mathcal{B}}{2}\right)
\end{equation*}
Witness that when $\eta=1/\mathcal{B}$, we can further simplify this to:
\begin{equation*}
    \mathcal{L}(\textit{\textbf{w}}^{t+1}) \lessapprox \mathcal{L}(\textit{\textbf{w}}^{t}) - \eta \norm{\nabla{\mathcal{L}(\textit{\textbf{w}}^t)}} ^2 
\end{equation*}
Since our NN satisfies the $\mu$-PL$_{*}$ condition from Eq.~\ref{muPL} due to its width, we therefore have: 
\begin{equation*}
    \mathcal{L}(\textit{\textbf{w}}^{t+1}) \leq (1- \eta \mu) \mathcal{L}(\textit{\textbf{w}}^{t}). 
\end{equation*}

\section{Hessian for Poisson's Equation} \label{poisson_stiff_dynamics}
The Hessian of the PDE part of the loss function is calculated following the methods introduced in \cite{wang2021understanding}, where the gradient of loss function is:
\begin{equation*}
    \pdv{\mathcal{L}_r}{w} = \pdv{\int_0^1 (\pdv[2]{u_w}{x} - \pdv[2]{u}{x})^2 dx}{w}.
\end{equation*}
Here, $u(x)$ is the target solution admitting some parameter $C$, and $u_w(x)$ is the NN approximation of the output. Assuming that the approximation is a good representation of the actual solution, it can be written as $u_w(x) = u(x)\epsilon_{w}(x)$, where $\epsilon_w(x)$ is a smooth function taking values in $[0,1]$, such that $|\epsilon_w(x)-1|<\delta$ for some $\delta>0$ and $\abs{\pdv[k]{\epsilon_w(x)}{x}} \leq \delta$.

The Hessian of the loss function will therefore be:
\begin{equation*}
    \pdv[2]{\mathcal{L}_r}{w} = \pdv[2]{\int_0^1 (\pdv[2]{u_w}{x} - \pdv[2]{u}{x})^2}{w} = \pdv{I}{w}
\end{equation*}
where $I = \pdv{\mathcal{L}_r}{w}$, and can be calculated to be: 
\begin{align}
I = 2 {\pdv[2]{u_w}{w}{x}} \left(\pdv[2]{u_w}{x} - \pdv[2]{u}{x}\right)\Big|_0^1 - {\pdv{u_w}{w}} \left(\pdv[3]{u_w}{x} - \pdv[3]{u}{x}\right)\Big|_0^1 + \int_0^1 {\pdv{u_w}{w}}  \left(\pdv[4]{u_w}{x} - \pdv[4]{u}{x}\right) dx.
\end{align}
The above equation contains 3 terms, which we call $I_1$, $I_2$, and $I_3$ respectively, and define $\pdv {I}{w} = \pdv {I_1}{w} + \pdv {I_2}{w} + \pdv {I_3}{w}$. We calculate these terms as follows:
\begin{align*}
I_1 &= 2 \pdv[2]{u_w}{w}{x} \left(\pdv[2]{u_w}{x} - \pdv[2]{u}{x}\right)\Big|_0^1 \\
\pdv{I_1}{w} &= 2 \pdv{({\pdv[2]{u_w}{w}{x}})}{w} (\pdv[2]{u_w}{x} - \pdv[2]{u}{x}) \left(\pdv[2]{u_w}{x} - \pdv[2]{u}{x}\right)\Big|_0^1\\
             &= 2 \left(\pdv{\pdv{(u'(x)\epsilon_w(x) + u(x) \epsilon'_w(x))}{w}}{w}\right) \left(\pdv[2]{u_w}{x} - \pdv[2]{u}{x}\right)\Big|_0^1 \\
             &= 2 \left(\pdv{(u'(x)\pdv{\epsilon_w(x)}{w} + u(x) \pdv{\epsilon'_w(x)}{w})}{w}\right) \left(\pdv[2]{u_w}{x} - \pdv[2]{u}{x}\right)\Big|_0^1\\
             &= 2 \left(u'(x)\pdv[2]{\epsilon_w(x)}{w} + u(x) \pdv[2]{\epsilon'_w(x)}{w}\right) \left(\pdv[2]{u_w}{x} - \pdv[2]{u}{x}\right)\Big|_0^1.
\end{align*}

We assume both $u''$ and $u$ are bounded on our domain, hence without loss of generality taking $|u(x)|\leq 1$ and $|u'(x)|\leq 1$. An application of the triangle inequality and chain rule then yields:
\begin{equation*}
    \abs {u'(x)\pdv[2]{\epsilon_w(x)}{w} + u(x) \pdv[2]{\epsilon'_w(x)}{w}} \leq C \norm{\pdv[2]{\epsilon_w(x)}{w}}+\norm{\pdv[2]{\epsilon'_w(x)}{w}}.
\end{equation*} 
Using the assumptions of $\epsilon_w(x)$ we get:
\begin{equation*}
    \pdv[2]{u_w}{x} - \pdv[2]{u}{x}  \leq (C^2+2)\delta.
\end{equation*}
Combing both of these, we see $\pdv{I_1}{w}= \mathcal{O}(C^3)$. Similarly, $\pdv{I_2}{w}=  \mathcal{O}(C^3)$ and  $\pdv{I_3}{w} = \mathcal{O}(C^4)$. 

\section{Gradient Flow for SGD with Momentum} \label{SGDM_flow}

Recall that as the NN becomes wider, the norm of the Hessian becomes smaller such that in the limit as $N \to \infty$ the norm of Hessian becomes $0$. One immediate consequence of small Hessian for a NN is that its output can be estimated by a linear function \cite{lee2019wide}. The output of a NN can thus be replaced by its first order Taylor expansion: 
\begin{equation*}
    u^{\text{lin}}_t(\textit{\textbf{w}}) \approx u(\textit{\textbf{w}})|_{\textit{\textbf{w}}_{0}} + (\textit{\textbf{w}} - \textit{\textbf{w}}_0) \nabla u(\textit{\textbf{w}})|_{\textit{\textbf{w}}_{0}}.
\end{equation*}
The update rule for SGDM can be written as \cite{du2019frontier}:
\begin{equation*}
    \textit{\textbf{w}}_{t+1} = \textit{\textbf{w}}_{t} + \alpha(\textit{\textbf{w}}_{t} - \textit{\textbf{w}}_{t-1}) - \eta \nabla_{w}\mathcal{L}(\textit{\textbf{w}}).
\end{equation*}
The discrete updates to the output of NN become (see Appendix~\ref{linear_updates}):
\begin{equation*}
     u^{\text{lin}}_{t+1} = u^{\text{lin}}_{t} + \alpha(u^{\text{lin}}_{t} - u^{\text{lin}}_{t-1}) - \eta \nabla_{w}\mathcal{L}(\textit{\textbf{w}}) \nabla_{w}u(\textit{\textbf{w}})|_{\textit{\textbf{w}}_{0}}
\label{descrete} 
\end{equation*}
In the rest of the paper, for simplicity, we have dropped the ``lin'' term.
The dynamics of SGDM is analogous to the equation of motion of a point mass $m$ undergoing a damped harmonic oscillation \cite{qian1999momentum}:
\begin{equation*}
    m\ddot{u} + \mu \dot{u} - \nabla \mathcal{L} f(u)= 0
\end{equation*}
where $f(u)$ is a linear function of $u$, and $\mu$ is the friction coefficient that is related to momentum: $\alpha= \frac{m}{m + \mu \Delta t}$ \cite{qian1999momentum}. Thus, the gradient flow of $u_{t}$ and $\textit{D}[{u}](x,w(t))$ can be written as: 
\begin{equation} 
\begin{split}
    m \ddot{u}(\textit{\textbf{x}}_{b},\textit{\textbf{w(t)}}) \,=& -\mu \dot{u}(\textit{\textbf{x}}_{b},\textit{\textbf{w(t)}}) - \textit{\textbf{K}}_{{bb}_{(x,x')}}(\textit{\textbf{w}})(u(\textit{\textbf{x}}_{r},\textit{\textbf{w(t)}}) - g(\textit{\textbf{x}}_{b})) \\
    &-  \textit{\textbf{K}}_{{rb}_{(x,x')}}(\textit{\textbf{w}}) (\textit{D}[u](\textit{\textbf{x}}_{r},\textit{\textbf{w(t)}}) - h(\textit{\textbf{x}}_{r}))\\
    m\textit{D}[\ddot{u}](\textit{\textbf{x}}_{r},\textit{\textbf{w(t)}}) \,=& -\mu \textit{D}[\dot{u}](\textit{\textbf{x}}_{r},\textit{\textbf{w(t)}}) - \textit{\textbf{K}}_{{br}_{(x,x')}}(\textit{\textbf{w}})(u(\textit{\textbf{x}}_{b},\textit{\textbf{w(t)}}) - g(\textit{\textbf{x}}_{b})) \\
    &-  \textit{\textbf{K}}_{{rr}_{(x,x')}}(\textit{\textbf{w}}) (\textit{D}[u](\textit{\textbf{x}}_{r},\textit{\textbf{w(t)}}) - h(\textit{\textbf{x}}_{r})).
\label{flow_step1}
\end{split}
\end{equation}
As mentioned earlier, in wide NNs if the last layer of the network is linear then the tanget kernels are static.  

We write the matrix format of Eq.~\ref{flow_step1} as follows:
\begin{equation}
m\begin{bmatrix}
\ddot{u}(\textit{\textbf{x}}_{b},\textit{\textbf{w(t)}})   
\\
\textit{D}[\ddot{u}](\textit{\textbf{x}}_{r},\textit{\textbf{w(t)}})
\end{bmatrix} = -\mu \begin{bmatrix}
\dot{u}(\textit{\textbf{x}}_{b},\textit{\textbf{w(t)}})   
\\
\textit{D}[\dot{u}](\textit{\textbf{x}}_{r},\textit{\textbf{w(t)}}) \end{bmatrix} -
\textit{\textbf{K}} \begin{bmatrix}
u(\textit{\textbf{x}}_{b},\textit{\textbf{w(t)}}) - g(\textit{\textbf{x}}_{b}) \\
\textit{D}[u](\textit{\textbf{x}}_{r},\textit{\textbf{w(t)}}) - h(\textit{\textbf{x}}_{r})
\end{bmatrix}
\label{matrix_format}
\end{equation}


As $\textit{\textbf{K}}$ is a positive semi-definite matrix \cite{wang2021understanding} Eq.~\ref{matrix_format} can be viewed as a set of independent differential equations, each one corresponding to an eigenvalue $\lambda_{i}$ of the kernel. These give rise to the individual general solutions of the form:
\begin{equation*}
\begin{split}
    A_1 e^{ \lambda_{{i}_1} t} +  A_2 e^{ \lambda_{{i}_2} t}\\
    \lambda_{{i}_{1,2}} = -\gamma \pm \sqrt{\gamma^2 - \kappa'_{i}}
\end{split}
\end{equation*}
where $A_1$ and $A_2$ are two constants, $\gamma = \mu/2$, and $\kappa'_{i} = \frac{\kappa_{i}}{m}$. Of note, $\mu$ and $m$ are set by the user, which in turn define the value of momentum. The choice of values of $m$ and $\mu$ will determine the rate of decay of the above equation.  




\section{Linear NN Updates} \label{linear_updates}

The Taylor expansions of the output at time steps $t+1$ and $t$ are:
\begin{align*}
    u^{\text{lin}}_{t+1}(\textit{\textbf{w}}) &= u(\textit{\textbf{w}})|_{\textit{\textbf{w}}_{0}} + (\textit{\textbf{w}}_{t+1} - \textit{\textbf{w}}_0) \nabla u(\textit{\textbf{w}})|_{\textit{\textbf{w}}_{0}}\\
    u^{\text{lin}}_{t}(\textit{\textbf{w}}) &= u(\textit{\textbf{w}})|_{\textit{\textbf{w}}_{0}} + (\textit{\textbf{w}}_{t} - \textit{\textbf{w}}_0) \nabla u(\textit{\textbf{w}})|_{\textit{\textbf{w}}_{0}}.
\end{align*}
Thus, the difference between the outputs in the interval is: 
\begin{align*}
u^{\text{lin}}_{t+1}(\textit{\textbf{w}}) - u^{\text{lin}}_{t}(\textit{\textbf{w}})  &= \nabla u(\textit{\textbf{w}})|_{\textit{\textbf{w}}_{0}} (\textit{\textbf{w}}_{t+1} - \textit{\textbf{w}}_{t})\\
    &=  \nabla u(\textit{\textbf{w}})|_{\textit{\textbf{w}}_{0}} (\alpha (\textit{\textbf{w}}_{t} - \textit{\textbf{w}}_{t-1}) - \eta \nabla \mathcal{L}(\textit{\textbf{w}}_{t}))
\end{align*}
where we used the update rule of SGD with momentum within the equation. Similarly, we have:
\begin{equation*}
    u^{\text{lin}}_{t}(\textit{\textbf{w}}) - u^{\text{lin}}_{t-1}(\textit{\textbf{w}}) = \nabla u(\textit{\textbf{w}})|_{\textit{\textbf{w}}_{0}} (\textit{\textbf{w}}_{t} - \textit{\textbf{w}}_{t-1})
\end{equation*}
and:
\begin{align*}
u^{\text{lin}}_{t+1}(\textit{\textbf{w}}) - u^{\text{lin}}_{t}(\textit{\textbf{w}}) =  \nabla u(\textit{\textbf{w}})|_{\textit{\textbf{w}}_{0}} \left(\alpha \left(\frac{u^{\text{lin}}_{t}(\textit{\textbf{w}}) - u^{\text{lin}}_{t-1}(\textit{\textbf{w}})}{\nabla u(\textit{\textbf{w}})|_{\textit{\textbf{w}}_{0}}}\right) - \eta \nabla \mathcal{L}(\textit{\textbf{w}}_{t})\right).
\end{align*}
Thus:
\begin{equation*}
u^{\text{lin}}_{t+1}(\textit{\textbf{w}}) = u^{\text{lin}}_{t}(\textit{\textbf{w}}) - \eta \nabla \mathcal{L}(\textit{\textbf{w}}_{t})\cdot\nabla u(\textit{\textbf{w}})|_{\textit{\textbf{w}}_{0}} - \alpha ({u^{\text{lin}}_{t}(\textit{\textbf{w}}) - u^{\text{lin}}_{t-1}(\textit{\textbf{w}})}).
\end{equation*}

\section{Experiment Plots} \label{experiment plots}
\setcounter{figure}{0}
\renewcommand\thefigure{C.\arabic{figure}} 
\begin{figure}[h]
\centering
\includegraphics[width=1\columnwidth]{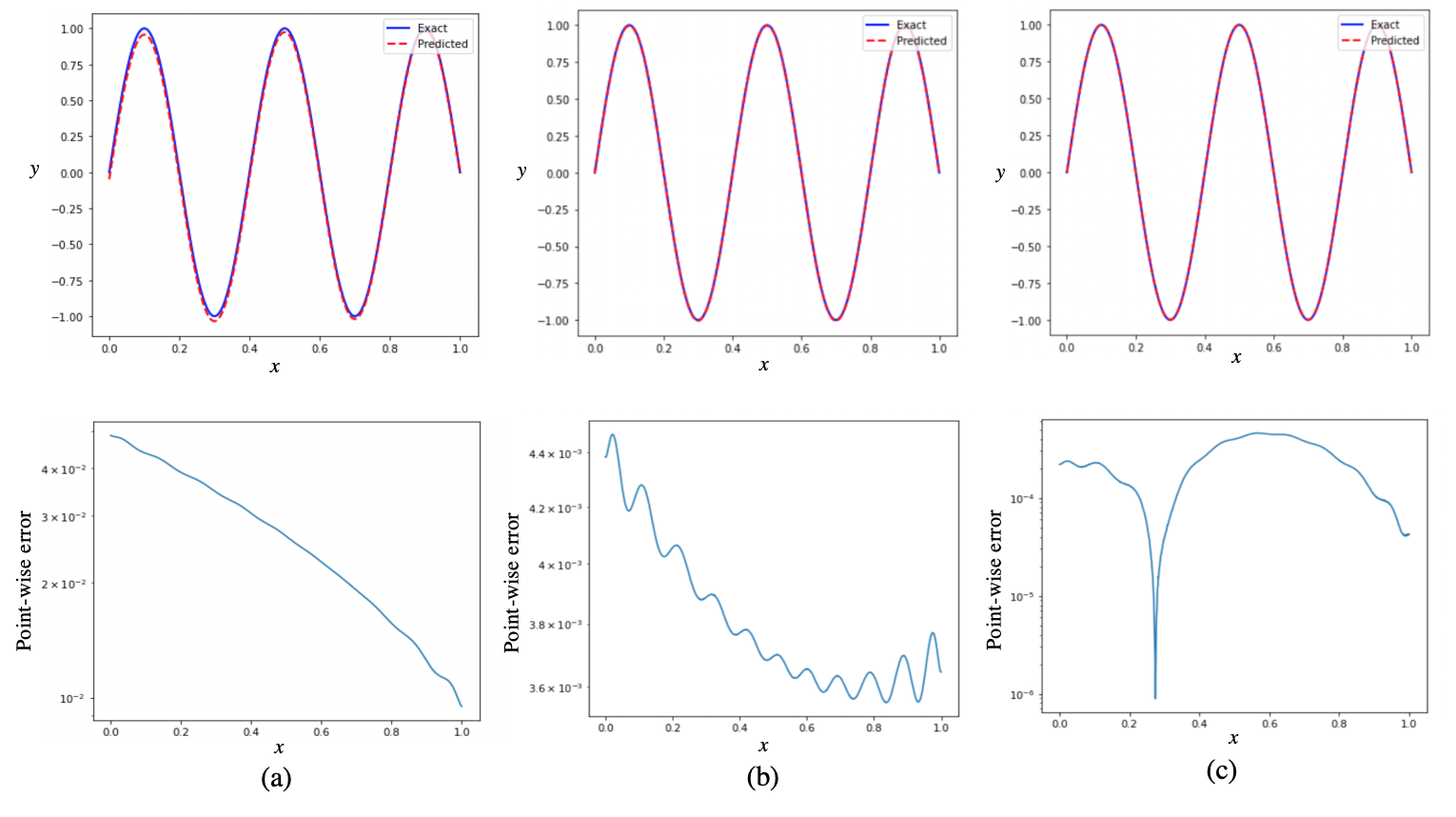}
\caption{1D Poisson equation when $C = 5 \pi$. The predicted solution is presented in red dashed line and the exact solution is plotted in solid blue line. The solutions are obtained by training a $1$-layer network with the width of 500. (a) Vanilla SGD after $45000$ epochs with relative error on order of $10^{-2}$(b) SGDM after $40000$ epochs with relative error on order of $10^{-3}$. (c) Adam after $15000$ epochs with relative error on order of $10^{-4}$.}
\label{fig:Poission_C5}
\end{figure} 
 
\begin{figure}[H]
\centering
\includegraphics[width=12cm, height = 4 cm]{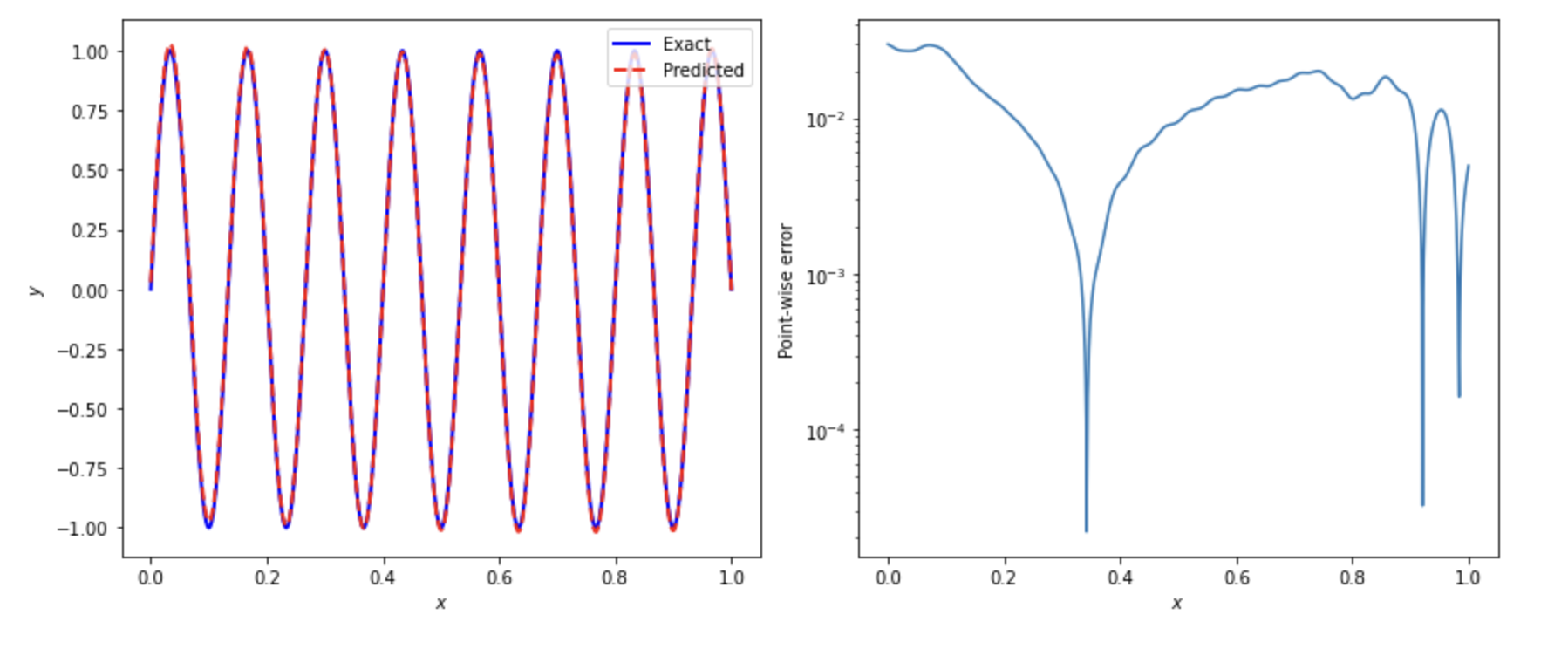}
\caption{The estimation of the solution of the Poisson's equation for $C = 15 \pi$, using the vanilla SGD algorithm after 225000 epochs. }
\label{fig:poission_18K_SGD}
\end{figure} 

\section{Comparison}

Here, we compare our numerical solution of the transport function, the reaction-diffusion problem and the Poisson's equation based on training the model using the Adam optimizer for a wide network, with the solutions presented in \cite{krishnapriyan2021characterizing} and \cite{wang2022}. In the transport function, $\beta = 40$ was the largest parameter that was analysed in \cite{krishnapriyan2021characterizing}, and in the reaction-diffusion problem, $\nu = 5$ and $\rho = 6$ were the largest shown parameters. Similarly, the solution of the Poisson's equation in \cite{wang2022} was shown when $C = 4\pi$. We used the same parameters to show the comparison between our methodology and the mentions studies. In Table~\ref{tab:method-comparion}, the relative and absolute errors of the estimated solutions (based on our approach and method presented in \cite{krishnapriyan2021characterizing} and \cite{wang2022}) are presented. Our solution for the transport function when $\beta = 40$ is as good as the presented solution in \cite{krishnapriyan2021characterizing} (the estimated solution from both methods have errors on order of $10^{-2}$). However, in the case of reaction-diffusion and the Poisson's equations our estimated solutions are more accurate and have smaller relative and absolute errors. Of note, we have already shown that for much larger choices of parameters (see Section~\ref{Sec:width}) our trained models are also capable of estimating accurate solutions.

To build our networks, we used the same number of hidden layers as used in \cite{krishnapriyan2021characterizing} and \cite{wang2022}. However, we made our networks much wider. Moreover, for Poisson's equation we trained our model, using the same number of epochs as proposed in \cite{wang2022}. The comparison between the choices of basic hyperparemters such as number of hidden layers, width of each layer, and number of epochs are shown in Table~\ref{tab:NAS-compare}.

\begin{table}[h]

\centering
\begin{tabular}{|l|l|l|l|}
\hline
 \textcolor{blue}{PDE of Interest}&
 \textcolor{blue} {Error Type} &
  \textcolor{blue}{Solution in \cite{krishnapriyan2021characterizing}}&
  \textcolor{blue}{Our Solution} \\ \hline
\begin{tabular}[c]{@{}l@{}}1D Convection \\ \\ $\beta$ = 40\end{tabular} &
  \begin{tabular}[c]{@{}l@{}}Relative Error\\ \\ Absolute error\end{tabular} &
  \begin{tabular}[c]{@{}l@{}}$5.33e-2$\\ \\$2.69e-2$\end{tabular} &
  \begin{tabular}[c]{@{}l@{}}$1.34e-2$\\ \\ $1.11e-2$\end{tabular} \\ \hline
\begin{tabular}[c]{@{}l@{}}1D reaction-diffusion\\ \\ $\nu$ = 5, $\rho$ = 6\end{tabular} &
  \begin{tabular}[c]{@{}l@{}}Relative Error\\ \\ Absolute Error\end{tabular} &
  \begin{tabular}[c]{@{}l@{}}$2.69e-2$\\ \\ $1.28e-2$\end{tabular} &
  \begin{tabular}[c]{@{}l@{}}$9.04e-3$\\ \\ $4.03e-3$\end{tabular} \\ \hline
  \hline
 \textcolor{blue}{PDE of Interest} &
  \textcolor{blue} {Error Type} &
  \textcolor{blue}{Solution in \cite{wang2022}}&
  \textcolor{blue}{Our Solution} \\ \hline
  \begin{tabular}[c]{@{}l@{}}1D Poisson's Equation \\ \\ C= 4$\pi$\end{tabular} &
  \begin{tabular}[c]{@{}l@{}}Relative Error\\ \\ Absolute error\end{tabular} &
  \begin{tabular}[c]{@{}l@{}}$1.63 e-3$\\ \\$3.80e-2$\end{tabular} &
  \begin{tabular}[c]{@{}l@{}}$6.95e-5$\\ \\ $4.45e-5$\end{tabular} \\ \hline
\end{tabular}
\caption{The relative and absolute errors of the solutions presented in \cite{krishnapriyan2021characterizing} and \cite{wang2022} for 1D transport function, the reaction-diffusion equation, and Poisson's equation are shown in the third column. The relative and absolute errors of the solutions presented using our methodology are also presented in the fourth column.}
\label{tab:method-comparion}
\end{table}

\begin{table}[h]
\scalebox{0.65}{
\centering
\begin{tabular}{|l|l|l|l|l|l|l|}
\hline
Neural Network Parameters &
\begin{tabular}[c]{@{}l@{}}Network of \cite{krishnapriyan2021characterizing}\\ for Transport Function\end{tabular} &
\begin{tabular}[c]{@{}l@{}}Our Network for\\ Transport Function\end{tabular} &
\begin{tabular}[c]{@{}l@{}}Network of \cite{krishnapriyan2021characterizing}\\ for reaction-diffusion\end{tabular} &
  \begin{tabular}[c]{@{}l@{}}Our Network for \\ reaction-diffusion\end{tabular} &
  \begin{tabular}[c]{@{}l@{}}Network of \cite{wang2022}\\ Poisson\end{tabular} &
  \begin{tabular}[c]{@{}l@{}}Our Network for\\ Poisson\end{tabular} \\ \hline
Number of  Hidden Layers & 4             & 4     & 4             & 4     & 1     & 1     \\ \hline
Width of Each Layer      & 50            & 500   & 50            & 500   & 100   & 500   \\ \hline
Epochs                   & Not Presented & 25000 & Not Presented & 55000 & 45000 & 45000 \\ \hline
\end{tabular}}
\caption{The neural network parameters such as number of hidden layers, width of each layer, and number of epochs are compared between our model and the presented models in \cite{krishnapriyan2021characterizing} and \cite{wang2022}.}
\label{tab:NAS-compare}
\end{table}

\end{document}